\documentclass[letterpaper, 10 pt, conference]{ieeeconf}  

\IEEEoverridecommandlockouts                              

\usepackage{amsmath,amssymb} \usepackage{url}
\usepackage{xcolor}
\usepackage{nccmath} 
\usepackage{float}
\usepackage{graphicx} 
\usepackage{gensymb} 
\usepackage{CJKutf8}
\usepackage[ruled,vlined,linesnumbered]{algorithm2e} 
\usepackage{color}
\usepackage{booktabs}
\usepackage{balance}
\usepackage{url}

\title{\LARGE \bf
Arm-Constrained Curriculum Learning for Loco-Manipulation of the Wheel-Legged Robot 
}

\author{Zifan Wang$^{*1}$, Yufei Jia$^{*3}$, Lu Shi$^{*2}$, Haoyu Wang$^{2,4}$, Haizhou Zhao$^{2,5}$, Xueyang Li$^{6}$, \\
Jinni Zhou$^{1\dag}$, Jun Ma$^{1\dag}$, and  Guyue Zhou$^{2\dag}$ 
\thanks{*Equal contribution. $^{1}$The Hong Kong University of Science and Technology (Guangzhou), $^{2}$Institute for AI Industry Research (AIR), Tsinghua University,  $^{3}$Department of Electronic Engineering, Tsinghua University, $^{4}$Harbin Institute of Technology, $^{5}$Xi'an Jiaotong-Liverpool University, $^{6}$DISCOVER Robotics}
\thanks{$\dag$Corresponding author. eejinni@hkust-gz.edu.cn, jun.ma@ust.hk, zhouguyue@air.tsinghua.edu.cn}%
}

\begin{document}

\maketitle
\thispagestyle{empty}
\pagestyle{empty}


\begin{abstract}
Incorporating a robotic manipulator into a wheel-legged robot enhances its agility and expands its potential for practical applications. However, the presence of potential instability and uncertainties presents additional challenges for control objectives. In this paper, we introduce an arm-constrained curriculum learning architecture to tackle the issues introduced by adding the manipulator. Firstly, we develop an arm-constrained reinforcement learning algorithm to ensure safety and stability in control performance. Additionally, to address discrepancies in reward settings between the arm and the base, we propose a reward-aware curriculum learning method. The policy is first trained in Isaac gym and transferred to the physical robot to do dynamic grasping tasks, including the door-opening task, fan-twitching task and the relay-baton-picking and following task. The results demonstrate that our proposed approach effectively controls the arm-equipped wheel-legged robot to master dynamic grasping skills, allowing it to chase and catch a moving object while in motion. Please refer to our website (\url{https://acodedog.github.io/wheel-legged-loco-manipulation/}) for the code and supplemental videos. 
\end{abstract}

\begin{figure*}[t]
\vspace{6pt}
    \centering
    \includegraphics[width = 0.9\textwidth]{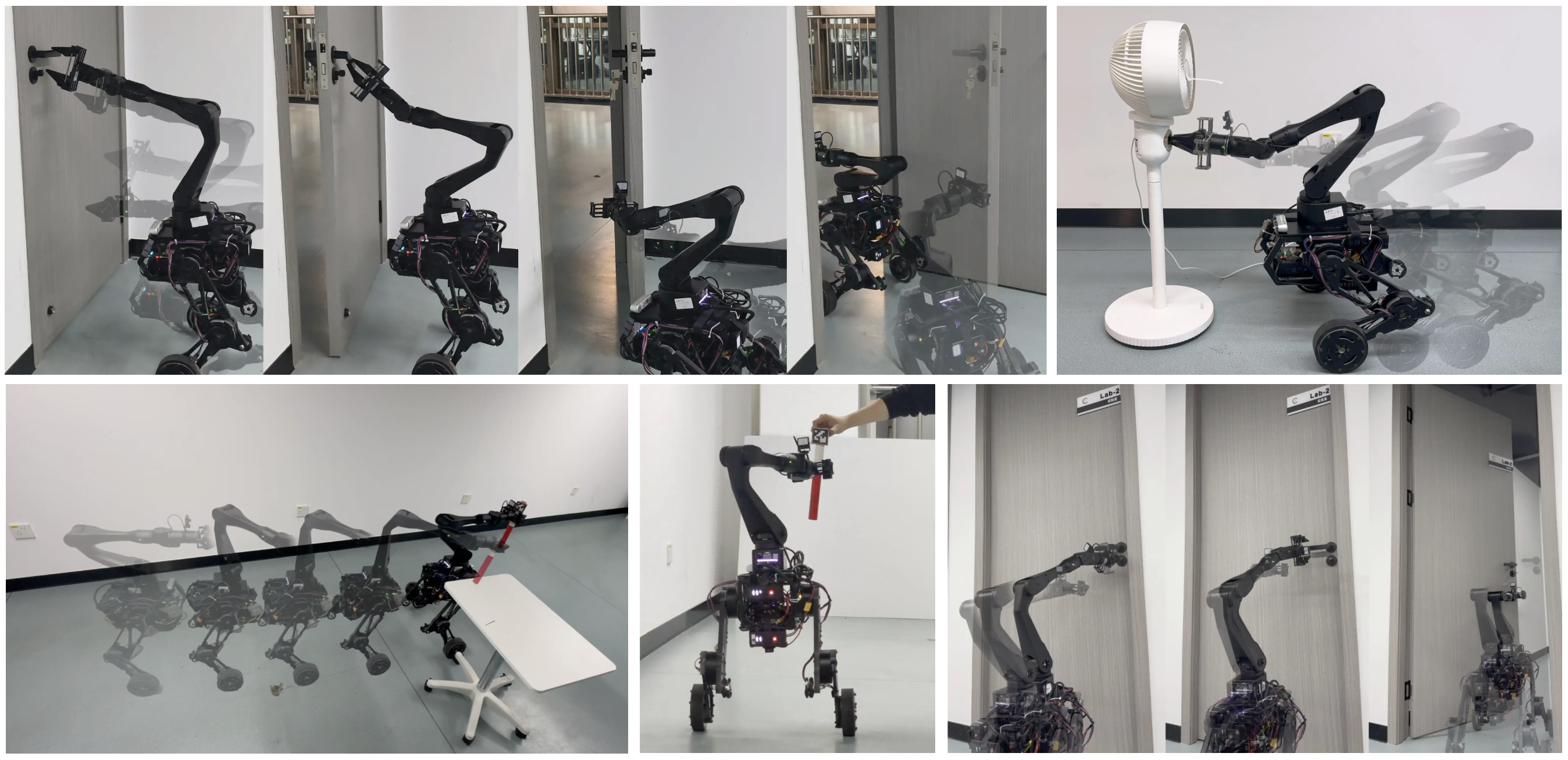}
    \caption{Tasks accomplished by the proposed architecture. Top-Left: door-opening-and-pulling task; Top-Right: fan-knob-twitching task; Bottom-Left: relay-baton-chasing task; Bottom-Right: door-opening-and-pushing task.}
    \vspace{-9pt}
    \label{fig:tasks}
    \vspace{-9pt}
\end{figure*}

\section{Introduction}
Human beings are capable of easily completing a variety of complex tasks, such as navigating through areas with various obstacles and interacting with different objects. However, these seemingly common tasks introduce many challenges for robots to master these skills: robots need to perform dynamic movements at high speeds and interact with different objects while coordinating their various parts to ensure safety. The emergence of legged robot platforms has provided a feasible foundation for executing complex tasks in various uncertain environments~\cite{jenelten2024complex,geilinger2020WL}. 

By augmenting robotic arms to the mobile robotic platform, e.g., quadrupedal robots~\cite{fu2023quadArm}, bipedal legged robots~\cite{raza2021bipadelArm}, and wheel-legged robots~\cite{chang2016leggedarm}, etc, more complex interactions can be achieved. Among them, wheel-legged robots leverage the benefits of both wheeled and legged robots, harnessing the high energy efficiency of wheels along with the superior adaptability to surmount uneven terrain and obstacles using legs~\cite{wang2021WLLQR}. Research on robot whole-body control and object manipulation using Deep Reinforcement Learning (DRL) is increasingly growing which provides a more sample-efficient, flexible, and robust strategy for learning, and enables the robot to more easily understand how to accomplish various complex tasks~\cite{morales2021survey,zhang2024asym}. 

Integrating a robot manipulator with a wheel-legged robot enhances agility and unlocks greater potential in practical applications. However, despite extensive studies on legged robots, a critical gap exists in the manipulation capabilities of wheel-legged robots. Coordinating the movement of both the arms and the wheels simultaneously requires managing multiple control modes, each with its own dynamics and constraints, that present additional challenges for the control. On the other hand, balancing the robot while moving its arms and wheels introduces extra dynamic stability requirements. In this work, we propose an arm-constrained curriculum RL framework for loco-manipulation of the wheel-legged robot. The game-inspired curriculum learning procedure enables the simultaneous control of both the arm and wheels. An arm-constrained network is introduced in the framework to ensure the safety and stability of the robot. Our novel framework enables dynamic grasping, leveraging the inherent stability, efficiency, and speed advantages of wheel-legged platforms, thereby enhancing their object manipulation capabilities.

In essence, the paper contributes to the following aspects:
\begin{itemize}
    \item We proposed an arm-constrained curriculum reinforcement learning framework specifically designed for loco-manipulation of wheel-legged robots. The framework allows for simultaneous control of both the arm and wheels, addressing the stability, safety, and efficiency challenges of coordinating hybrid locomotion and manipulation tasks.
    \item We introduced a reward-aware curriculum learning process aimed at fostering balanced progress across all components of the agent, regardless of whether they have sparse or dense rewards. By implementing this approach, the risk of the system becoming stuck in a local minimum is mitigated.
\end{itemize}





\section{Related work}

\subsection{RL-based Control of Legged Robots}
Recent studies have shown impressive control performance when utilizing RL for legged robotic systems~\cite{LearningtoWalk2021},~\cite{9028188},~\cite{lee2020learning},~\cite{hwangbo2019learning},~\cite{yang2020data}. In~\cite{schneider2024distri}, distributional RL is employed to train a risk-aware algorithm for the quadrupedal-legged robot, allowing it to dynamically adapt its behavior to different types of terrain. Meanwhile, researchers in~\cite{Learningrobustperceptive822} integrate proprioceptive data with noisy exteroceptive information to enable fast dynamic walking on various terrains. However, achieving such exceptional controllers with natural motion styles and high task performance often requires meticulous reward shaping to attain the desired behavior~\cite{deformableterrainscirobotics2256}. The researcher addresses this issue by employing motion imitation techniques, training control policies for simulated characters to replicate pre-recorded movements observed in animals or humans~\cite{RoboImitationPeng20}. During the training process, the use of rewards encourages the policy to imitate movements from a motion clip, eliminating the need for extensive reward adjustments.



\subsection{Loco-Manipulation for Legged Robots}
Utilizing a versatile robotic platform equipped with arms enables dynamic object manipulation which is a challenging task~\cite{farnioli2016toward}~\cite{wu2019teleoperation}~\cite{10160582}. In~\cite{Versatilescirobotics}, an innovative approach is proposed that simultaneously identifies comprehensive whole-body trajectories and contact sequences to tackle diverse loco-manipulation scenarios within predefined environments. This method integrates trajectory optimization, informed graph search, and sampling-based planning, resulting in emergent behaviors for a quadrupedal mobile manipulator capable of both prehensile and nonprehensile interactions, enabling it to perform real-world tasks. In another study by~\cite{ma2022combining}, a novel approach is developed that combines learning-based locomotion policies with model-based manipulation. This enables legged mobile manipulators to adapt to challenging terrains and achieve robust locomotion. Additionally,~\cite{fu2023quadArm}, proposes a novel method to enhance the versatility of legged robots by learning a single unified policy. This policy facilitates seamless coordination between manipulation tasks (using an attached arm) and locomotion, overcoming the limitations of traditional hierarchical control pipelines. The method leverages reinforcement learning, Regularized Online Adaptation, and Advantage Mixing to bridge the Sim2Real gap and achieve dynamic and agile behaviors across various task setups.


\section{Preliminary}
In this section, we will introduce the technical preliminaries of the work.

 \begin{figure*}[hpt!]
 \vspace{6pt}
    \centering
    \includegraphics[width = 0.95\textwidth]{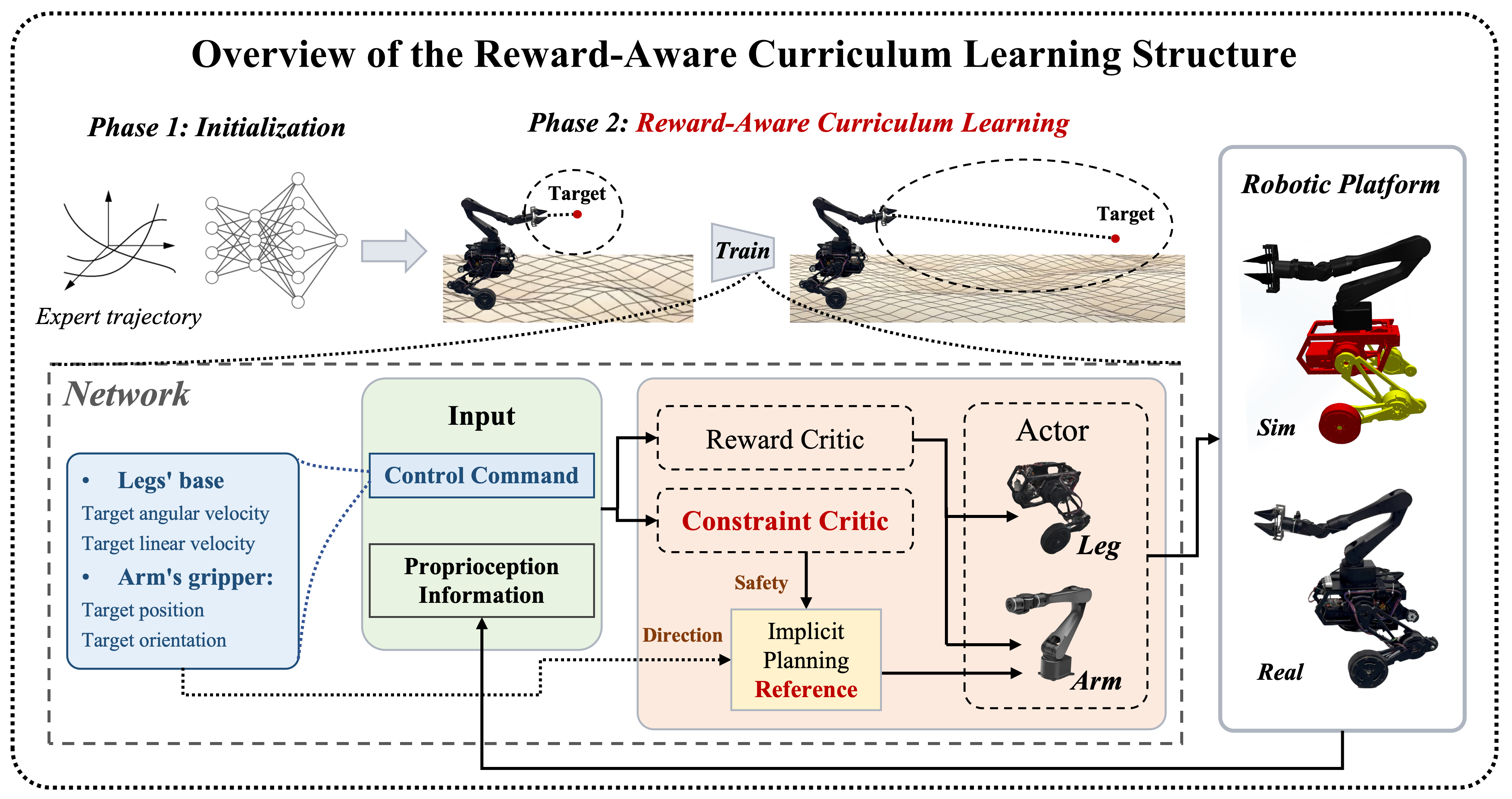}
    \caption{The overall illustration of the proposed framework. Top: two-phase learning procedure; Bottom: the detailed representation of the network.  
    }
    \vspace{-9pt}
    \label{fig:SystemFlow}
    \vspace{-9pt}
\end{figure*}

\subsection{Constrained Markov Decision Process (CMDP)}
The environment of an RL problem is typically stated in the form of a Markov Decision Process, which is a mathematical framework used to model decision-making problems. It is described by a tuple$(S, A, R, P, \rho,\gamma)$, where $S$ denotes the state space, $A$ denotes the action space,$ R \in R:S\times A \rightarrow \mathbb{R}$ is the immediate reward function, $P:S\times A\times S \rightarrow \mathbb{R}$ is the transition model, and $\gamma$ represents the discount factor. 
A Constrained Markov Decision Process (CMDP)~\cite{altman2021constrained}, is a variant of the traditional MDP framework where additional constraints are imposed on the state-action pairs or policies to satisfy specific requirements or limitations during decision-making~\cite{kim2023not},~\cite{lee2023evaluation}. It is usually described by a tuple $(S, A, R, P, C_{1,.., K}, \rho,\gamma)$, with the additional parameters $C_k:S\times A \rightarrow \mathbb{R}$ representing the cost function for $\forall k\in\{1,..., K\}$, and $\rho$ being the initial state distribution. To solve a CMDP, we aim to find a policy $\pi$ that maximizes:
\begin{equation}
	\label{eq_return}
 \begin{gathered}
	J_R(\pi) = {\mathbb{E}}\left[\sum_{t=0}^{\infty} \gamma^t r(s_t,a_t,s_{t+1})\right]\\
        \text{s.t. } J_{C_k} (\pi) \leq d_k \;\;\; \forall k \in \{1, ..., K\} \enspace ,\\
\end{gathered}
 \end{equation}
where $J_{C_k}$ is the constraint, $d_k$ is the constraint threshold, and the expectation $\mathbb{E}[\ldots]$ represents discounted expected return. 

\subsection{Constrained Proximal Policy Optimization}

Proximal Policy Optimization (PPO) is a model-free reinforcement learning algorithm that aims to find an optimal policy by iteratively updating a surrogate objective function while enforcing a constraint on the size of policy updates, leading to stable and efficient learning. The objective function is described as

\begin{equation*}
\begin{medsize}
L^{\text{PPO}}(\theta) = \underset{\substack{s \sim d^{\pi_i} \\
a \sim \pi}}{\mathbb{E}}\left[\frac{\pi_{\theta}(a \mid s)}{\pi_{i}(a \mid s)}A^{\pi_i}(s, a)\right] -\beta D_{K L}\left[\pi_{\theta}(\cdot \mid s), \pi_i(\cdot \mid s)\right] \enspace ,
\end{medsize}
\end{equation*}
where $D_{K L}$ is Kullback-Leibler Divergence and $\beta$ represents weight of the strategy.

Penalized PPO (P3O)~\cite{zhang2022penalized} is a variant of the PPO algorithm where additional constraints are incorporated into the optimization process to enforce specific criteria or limitations on the learned policies, ensuring compliance with desired behavior or safety constraints during training. The optimization problem becomes
\begin{equation}\label{eq:cppo}
\begin{medsize}
\begin{aligned}
    \underset{\pi \in \Pi_\theta}{\mathrm{maximize}}~L^{\text{P3O}}= L^{\text{PPO}}(\theta)  +\sum_{k=1}^K \log \left(d_k-J_{C_k}\left(\pi_{\theta}\right)\right) / t\\
    J_{C_k}\left(\pi_{\theta}\right)=J_{C_k}\left(\pi_i\right)+\frac{1}{1-\gamma} \underset{\substack{s \sim d^{\pi_i} \\
a \sim \pi}}{\mathbb{E}}\left[\frac{\pi_{\theta}(a \mid s)}{\pi_{i}(a \mid s)}A_{C_k}^{\pi_i}(s, a)\right]\enspace,
\end{aligned}
\end{medsize}
\end{equation}
where $J_{C_k}$ represents the constraint function introduced by the additional constraints $C_k, k\in{1,\dots,K}$.

\section{Method}\label{Sec:Method}
\subsection{Overview of the structure}
The overview of the proposed architecture is depicted in Fig.~\ref{fig:SystemFlow}. The entire structure follows a two-phase learning procedure. Initially, a behavioral cloning process is employed to initialize the actor network. Subsequently, a reward-aware curriculum learning process is executed to iteratively improve the policy for task completion. Specifically, the network is trained to follow user commands, which comprise two components: the desired target pose for the arm and the desired velocities for the body.  These user commands, along with proprioception information from the robot, serve as the input (observation) to the Arm-Constrained Proximal Policy Optimization (AC-PPO) policy networks. The AC-PPO framework consists of three networks: the actor network, the reward critic network, and the additional constraint critic network. Safety information generated by the constraint critic network is incorporated into the implicit planning reference used to guide the robot arm. This framework enables coordinated hybrid locomotion and manipulation tasks. The details of each component are introduced as follows.

\subsection{Arm-Constrained Proximal Policy Optimization}
One of the most challenging aspects of loco-manipulation for wheel-legged robots involves effectively coordinating locomotion and manipulation while adapting to dynamic objects. To ensure that the manipulation task does not compromise locomotion safety, we formulate the velocity-tracking and loco-manipulation tasks as a CMDP. Addressing the aforementioned challenge, we incorporate designed constraints of the arm in~\eqref{eq:cppo} and train the P3O algorithm to generate stable and safe actions for \textit{both the arm and the leg}. Specifically, the constraints are designed as follows:


\begin{enumerate}
    \item \textbf{Arm joint constraints:} Although the action space may impose some constraints on the values of the action, these limits cannot always be reached in a continuous action space. Additionally, we aim to avoid situations where the arm's joint positions consistently reach their limits, which could do harm to the motors of the robot. To achieve this, we impose joint position constraints to encourage actions that stay as far as possible from the limits. This helps reduce the occurrence of unsafe events and unstable movements. The expression is:
        \begin{equation*} 
        \begin{aligned}
            C_{\text{arm}} &= \sum_{i\in\text{arm joints}} \lvert\lvert\max(q_{i,t} - q^{\text{upper}}_i,0)\rvert\rvert_1 \\ 
            & + \sum_{i\in\text{arm joints}} \lvert\lvert\min(q_{i,t} - q^{\text{lower}}_i,0)\rvert\rvert_1\enspace .
            \end{aligned}
        \end{equation*}
    \item \textbf{Gripper position constraints:} This constraint pertains to the separation distance between the arm centroid
 and its base centroid. Adjusting the gripper’s position can alter the overall center of gravity of the robot, potentially leading to instability. To mitigate the risk of tipping over during manipulation tasks, this constraint is imposed to safeguard the robot’s stability and ensure its safety. The expression is described as:
    \begin{equation*} 
            C_{\text{Gripper}} = \lvert\lvert(P_{\text{centroid}}^{\text{arm}} - P_{\text{centroid}}^{\text{base}})\rvert\rvert_2 \enspace .
        \end{equation*}
     \item \textbf{Collision constraints:} To prevent self-collision and collisions between the robot arm and the environment, we impose a collision constraint on the force $f_i$ exerted on each arm link. Ideally, this force should remain close to 0 when no collision is detected, indicating a balanced state. Enforcing this constraint is essential to ensure the robot operates safely within its environment. We present the constraint as: 
    \begin{equation*} 
            C_{\text{force}} =  \sum_{i\in\text{arm links}}  \lvert\lvert(f_{i}^{\text{arm}})\rvert\rvert_2  \enspace .
        \end{equation*}
\end{enumerate}

\subsection{Reward-Aware Curriculum Learning}\label{subsec:curri}
Previous research has highlighted the advantages of employing a curriculum of task difficulty to train complex policies~\cite{xie2020curriculum},~\cite{bengio2009curriculum}. The fundamental concept involves initially training the policy on simpler tasks before gradually increasing the complexity. In our framework, we introduce a reward-aware curriculum learning approach.  Instead of adjusting the complexity of the tasks to achieve the curriculum, we initialize the agent closer to the target pose and progressively expand the range. This is especially beneficial for addressing the challenge of sparse reward settings, as the agent can quickly achieve a high reward in the early stages of the training process. In complex robotic platforms where the workspace and reward settings vary significantly across different components of the agent, we can encourage parts of the agent with sparse rewards to progress more equally alongside parts with dense rewards by employing this reward-aware curriculum learning process. 

In this specific scenario, the robot arm has a broader workspace but receives sparser rewards compared to the wheeled-legs. Consequently, the robot may become trapped in a local minimum where the legs receive high rewards, leading to conservative actions of the arm to avoid affecting the stability of the legs. To overcome this challenge and enhance training efficiency, this reward-based curriculum learning structure is deployed to encourage the arm to learn to move. We initialize the end-effector of the arm close to the goal position, enabling relatively easy movements to achieve high rewards at the beginning of the training. Consequently, it has less possibility to resist movement as we further increase the distance to the goal. An example of the procedure is: 
\begin{equation}\label{eq:represent}
\begin{gathered}
	P_{\text{ee}}^{\text{init}} \sim N(\mu,\Sigma) \\
        \text{s.t. }  \mu = P^{\text{goal}}, \enspace \Sigma = 1+\frac{t}{T} *D^2 \enspace ,\\
\end{gathered}
\end{equation}
where $P_{\text{ee}}^{\text{init}}$ is the gripper position where the robot is reset each time. $\mu$ and $\Sigma$ are the mean and variance of the Gaussian distribution. $P^{\text{goal}}$ is the target location of the gripper. $T$ is the maximum episode length and $D$ is the maximum desired target distance. Note that the expression of the process can be easily extended to other types of distributions. This approach enables both the legs and the manipulator to learn to track the references effectively.

\subsection{Two-phase Learning using Behavior Cloning}
As depicted in Fig.~\ref{fig:SystemFlow}, a two-phase learning process was implemented to ensure the safety and efficiency of the architecture. In the first phase, data is collected from a Whole-Body Controller~\cite{todorov2012mujoco} operating within the target environment. Utilizing the baseline controller ensures the safe generation of data without endangering the robot. Subsequently, this dataset is employed to initialize the parameters of the actor-network in the subsequent phase, utilizing behavioral cloning techniques to replicate the decisions made by the baseline controller.





\section{Experiments}
\subsection{Experimental Setup}
\begin{figure}[h]
    \centering
    \includegraphics[width = 0.48\textwidth]{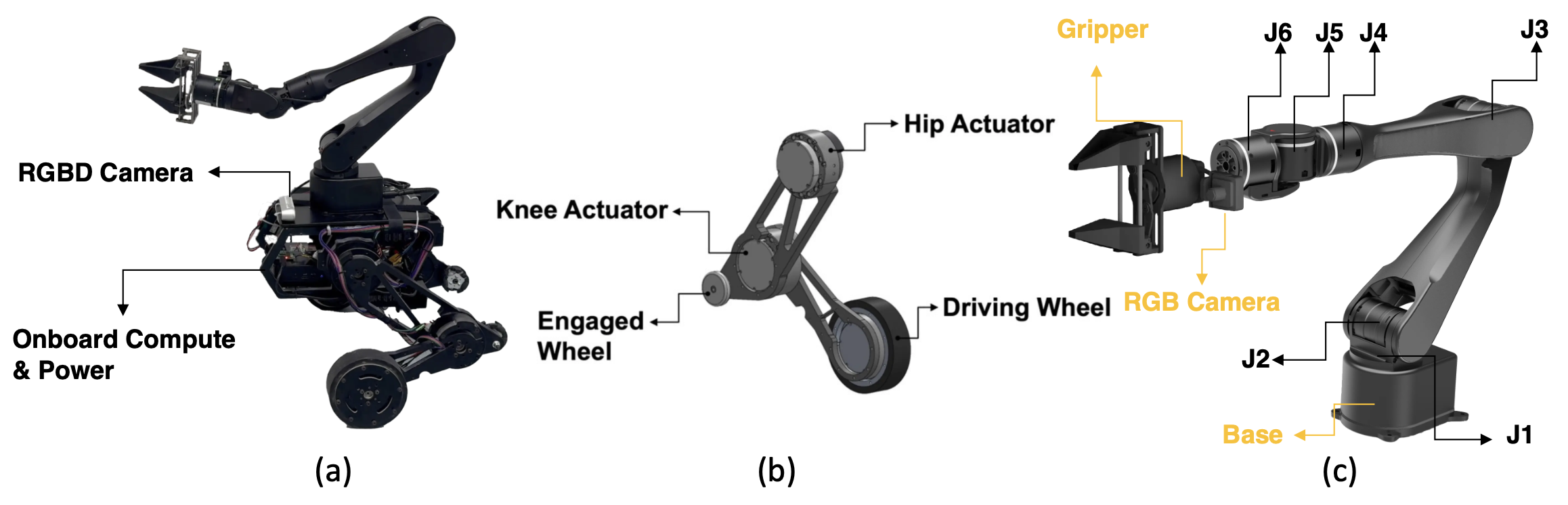}
    \caption{Illustration of the arm-equipped wheel-legged robotic platform.}
    \vspace{-6pt}
    \label{fig:robot}
    \vspace{-6pt}
\end{figure}

Several experiments are conducted to test the performance and stability of an arm-equipped wheel-legged robot as shown in Fig.~\ref{fig:robot}, which consists of a wheel-legged chassis and a robot arm. For each leg, there are a total of 3 motors: a hip motor, a knee actuator, and a driving wheel. Additionally, it's equipped with a 6-DoF serial robotic arm AIRBOT-Play and a gripper as the end-effector. Two cameras are mounted respectively on the robot's base and the end effector of the robot arm. Power is provided by onboard battery and computation is also done onboard. 

\begin{figure*}[htp!]
\vspace{6pt}
    \centering
    \includegraphics[width = 0.98\textwidth]{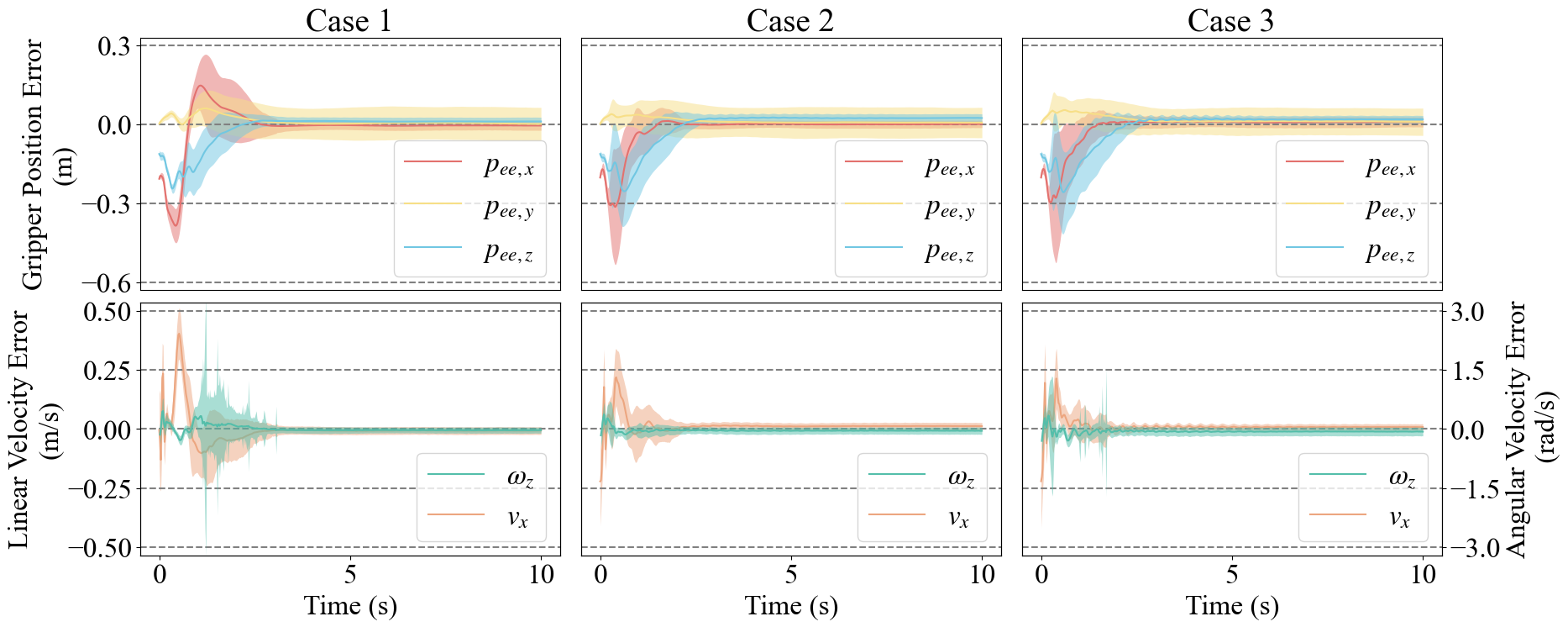}
    \caption{The tracking results in simulation. }
    \vspace{-9pt}
    \label{fig:exp-sim}
    \vspace{-6pt}
\end{figure*}

To train the policy according to the architecture outlined in Sec.~\ref{Sec:Method}, we defined the observation space, action space, and reward as follows:

    \subsubsection{\textbf{Observation Space}} The observation space is constructed by the state vector $S_t = \{S_t^{\text{base}}, S_t^{\text{arm}},S^{\text{cmd}}_t\}\in\mathbb{R}^{46}$. The information of the base is represented in $S_t^{\text{base}}=[h,v,\omega,R, q_{\text{leg}},\dot{q}_{\text{leg}},\dot{q}_{\text{wheel}}]\in\mathbb{R}^{20}$. The observation of arm is stored in $S_t^{\text{arm}}=\{q_{\text{arm}},\dot{q}_{\text{arm}},p_{\text{ee}},R_{\text{ee}}\}\in\mathbb{R}^{18}$, and $S^{\text{cmd}}_t=\{v_x^{\text{cmd}},\omega_{z}^{\text{cmd}},{p}_{\text{ee}}^{\text{cmd}},{R}_{\text{ee}}^{\text{cmd}}\}\in\mathbb{R}^{8}$ includes the control commands. The meanings of each parameter are:   
    \begin{itemize}
        \item $h \in\mathbb{R}^{1}$: base height.
        \item $v \in\mathbb{R}^{3}$: base linear velocity.
        \item $\omega \in\mathbb{R}^{3}$: base angular velocity.
        \item $R \in\mathbb{R}^{3}$: base orientation.
        \item $q_{\text{leg}} \in\mathbb{R}^{4}$: leg joints position.
        \item $\dot{q}_{\text{leg}} \in\mathbb{R}^{4}$: leg joints velocity.
        \item $\dot{q}_{\text{wheel}}\in\mathbb{R}^{2}$: wheel joints velocity.
        \item $q_{\text{arm}} \in\mathbb{R}^{6}$: arm joints position.
        \item $\dot q_{\text{arm}}\in\mathbb{R}^{6}$: arm joints velocity.
        \item $p_{\text{ee}}\in\mathbb{R}^{3}$: gripper position.
        \item $R_{\text{ee}}\in\mathbb{R}^{3}$: gripper orientation.
        \item $v_x^{\text{cmd}}\in\mathbb{R}^{1}$: x-axis linear velocity.
        \item $\omega_z^{\text{cmd}}\in\mathbb{R}^{1}$: z-axis angular velocity.
        \item ${p}_{\text{ee}}^{\text{cmd}}\in\mathbb{R}^{3}$: gripper target position.
        \item ${R}_{\text{ee}}^{\text{cmd}}\in\mathbb{R}^{3}$: gripper target orientation.
    \end{itemize}

 

\subsubsection{\textbf{Action Space}}
The action vector $A =\{ A_{\tau}, A_{\text{arm}} \}\in\mathbb{R}^{12}$ of the policy is constructed by the joint torques $A_{\tau}\in\mathbb{R}^{6}$ to control the wheeled legs and the joint position $A_{\text{arm}}\in\mathbb{R}^{6}$  to drive the arm. The vector of joint positions is further processed by a low-level controller to obtain the torque required to drive the arm.

\subsubsection{\textbf{Reward}}
As shown in Tab.~\ref{tab:rewards}, the locomotion rewards and manipulation rewards are defined separately for wheeled legs and the arm, both considered the tracking performance and safety.
%



\begin{table}[!htbp]
\vspace{-9pt}
\caption{Reward Function Settings}
\centering
\begin{tabular}{l|l}
\toprule
\multicolumn{1}{c|}{\textbf{Locomotion Rewards}} &  \multicolumn{1}{c}{\textbf{Expression}}\\ 
\hline
Linear velocity tracking & $\exp(-7.5 \cdot \lvert\lvert {v_x^{\text{cmd}}} - v_x\rvert\rvert^2$)\\
Angular velocity tracking & $ \exp(-1.25 \cdot \lvert\lvert {\omega^{\text{cmd}}_{z}} - \omega_{z}\rvert\rvert^2$)\\
Acceleration limits & $ -0.1 \cdot\lvert\lvert {v_x^{\text{last}}} - v_x\rvert\rvert^2$\\
Orientation  penalty& $ -1.2 \cdot\lvert\lvert {R_y}\rvert\rvert^2 -1.2 \cdot\lvert\lvert {R_x}\rvert\rvert^2$\\
Energy penalty&  $ -10^{-5} \cdot\lvert\lvert {\tau}\rvert\rvert^2$ , $\tau$: motor torque\\
Leg motion &$-10^{-7}( \ \lvert\lvert \dot{q}_{\text{leg}} \rvert \rvert ^2 -2.5 \ \lvert\lvert \ddot{q}_{\text{leg}} \rvert \rvert ^2 )$ \\
\midrule
\midrule
\multicolumn{1}{c|}{\textbf{Manipulation Rewards}} & \multicolumn{1}{c}{\textbf{Expression}}\\ 
\hline
Gripper position tracking & $ \exp(-5 \cdot \lvert\lvert {{p}_{\text{ee}}} - {p}_{\text{ee}}^{\text{cmd}}\rvert\rvert^2)$\\
Body position tracking & $ \exp(-0.05 \cdot \lvert\lvert {{p}_{\text{base}}} - {p}_{\text{base}}^{\text{cmd}}\rvert\rvert^2$)\\
Arm position upper limits & $-10 \ \sum \max(q_{\text{arm},i,t} - q^{\text{upper}}_{\text{arm},i},0)^2$ \\
Arm position lower limits & $-10 \ \sum \min(q_{\text{arm},i,t} - q^{\text{lower}}_{\text{arm},i},0)^2$ \\



\bottomrule
\multicolumn{2}{c}{}\\ 
\end{tabular}
\label{tab:rewards}
\vspace{-18pt}
\end{table}

\subsection{Simulation Tests}
The proposed structure is deployed in simulation first. We use Isaac Gym as our simulator and train the policy with $6000$ environments simultaneously. The control commands for the body, target linear and angular velocities of the base, are selected from uniform distributions over the intervals $v_x^{\text{cmd}} = [-2, 2]$m/s, and $\omega_z^{\text{cmd}} = [-0.5, 0.5]$rad/s at the beginning of the training. The robot then learned to track this velocity until termination. Upon meeting the termination criteria, which means the robot learned to track the current velocities, the target values were resampled. In terms of the manipulator, we employed the proposed reward-aware curriculum learning process as introduced in Sec.~\ref{subsec:curri}. The target position for the gripper is sampled from the normal distribution with $0$-mean and standard deviations within a narrow range initially, which is $[0.5, 0.1,0.2]$m for $p_{ee}^{\text{cmd}}$. Once the reward exceeds $90\%$ of the maximum reward, the range of standard deviations is expanded and the expanding step size is $[0.5, 0.1, 0.1]$m for the x-, y-, and z- directions of the end-effector.

Before transferring to the physical robot, we validate the performance of the algorithm in the simulation. Three cases of command velocities for the base are selected as $\{\text{Case 1: }v_x^{\text{cmd}} = 0.5$m/s, $\omega_z^{\text{cmd}} = 0$rad/s$\}$,
$\{\text{Case 2: }v_x^{\text{cmd}} = 0$m/s, $\omega_z^{\text{cmd}} = 0.5$rad/s$\}$, and $\{\text{Case 3: }v_x^{\text{cmd}} = 0.5$m/s, $\omega_z^{\text{cmd}} = 0.5$rad/s$\}$. They keep constant during the test procedure. On the other hand, the initial values of the desired target position for the manipulator are randomly selected from a cube of size $[0.25\times 0.2\times 0.35]\text{m}^3$ with the origin at $[0.25,0,0.15]$m. Then the initial target is propagated randomly within this range with a relatively small step size to produce a continuous target trajectory with a length of $500$. The sampling rate is $50$hz and the test is run repeatedly for $2024$ times to avoid bias. The test is designed to mimic the scenario that the robot tracks a dynamic target while moving its base. As shown in Fig.~\ref{fig:exp-sim}, our approach can track the command with a small error. Even for a moving target, it can converge quickly to the desired trajectory.

\subsection{Real-Robot Tests}
We further validate the trained policy by testing it on the physical robot to complete various tasks, as illustrated in Fig.~\ref{fig:tasks}. In contrast to the random generation during the simulation process, the control commands required for the base velocities and manipulation arm poses are provided through user inputs to accomplish different tasks. Based on the sources of these commands, our experiments can be categorized as follows:

\begin{enumerate}
\item \textbf{Teleoperation}:
The control commands are provided using a remote control handle. The robot can track these instructions to accomplish the following tasks.
    \begin{enumerate}
        \item \textit{Door-opening task:} This task involves approaching the doorknob, rotating it, pushing or pulling the door, and moving.
        \item \textit{Fan-twitching task:} This task involves approaching the fan, pushing and twitching the knob.        
    \end{enumerate}
    \item \textbf{Dynamic Manipulation}: Other than commands generated with human-in-the-loop, the challenging task of dynamic tracking is accomplished by obtaining the commands calculated from feedback provided by a camera. An RGB camera is installed on the gripper at the end of the robotic arm to guide the robot towards the relay baton and capture it. The robotic arm swiftly tracks the movement of the relay baton and ultimately grasps it using the proposed structure, highlighting the benefits of our algorithm in integrating the robot's chassis mobility with the dynamic movement of the robotic arm.


\end{enumerate}

\subsection{Ablation Study}
To evaluate the importance of different factors of our proposed structure, we did the ablation tests on two of the most important components in our method. 

\subsubsection{With/Without the Curriculum Learning}
\begin{figure}[H]
    \centering
   \includegraphics[width=\linewidth]{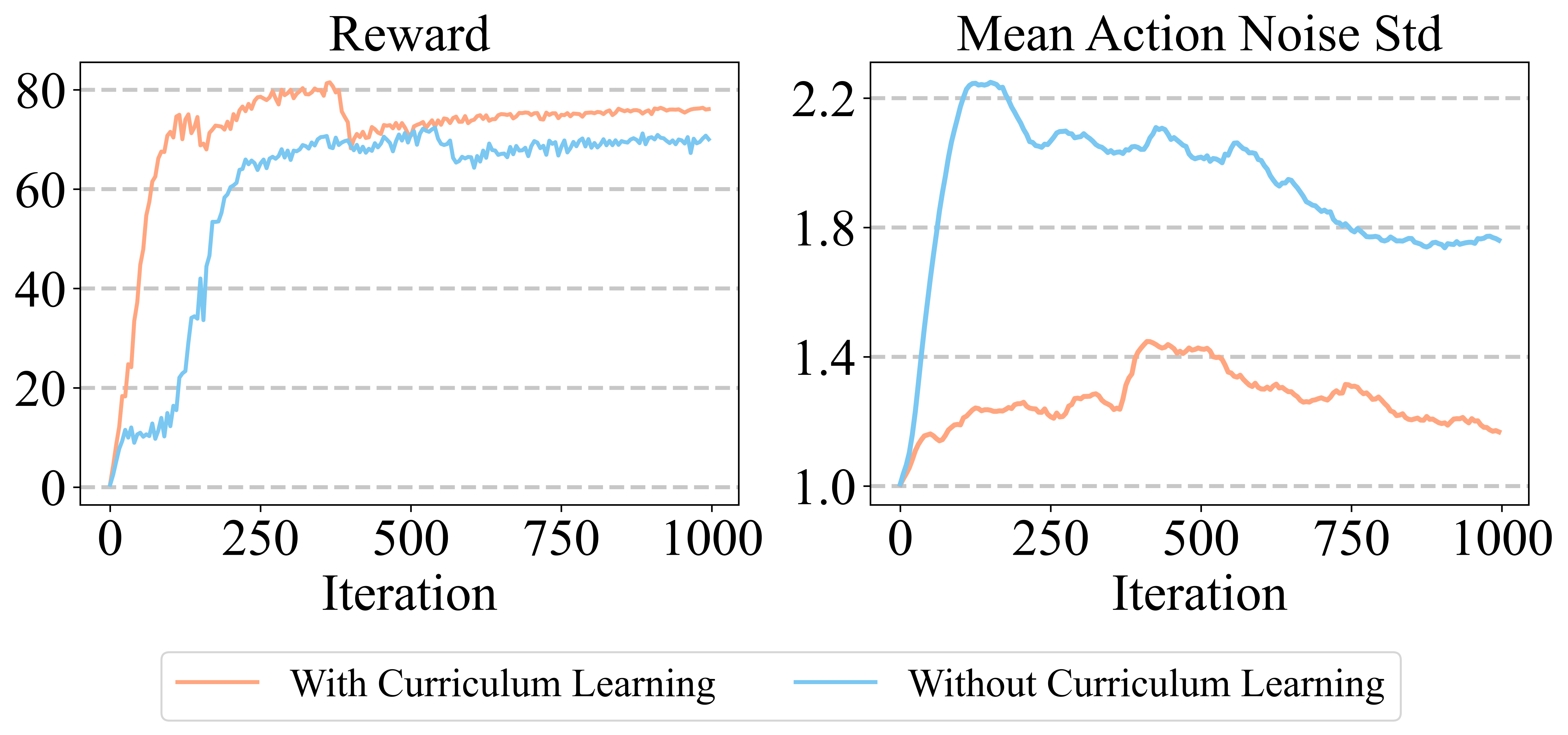}       
    \caption{Effect of curriculum learning for rewards and actions}
    \label{fig:CurriculumLearning}
\end{figure}

Fig.~\ref{fig:CurriculumLearning} illustrates the reward and mean action noise standard deviation throughout the training process. It is evident that our framework, incorporating the proposed reward-aware curriculum learning, achieves higher rewards and exhibits reduced randomness in actions.

\subsubsection{With/without the Arm-Constrained Critic Network}
\begin{figure}[H]
\vspace{6pt}
    \centering
   \includegraphics[width=\linewidth]{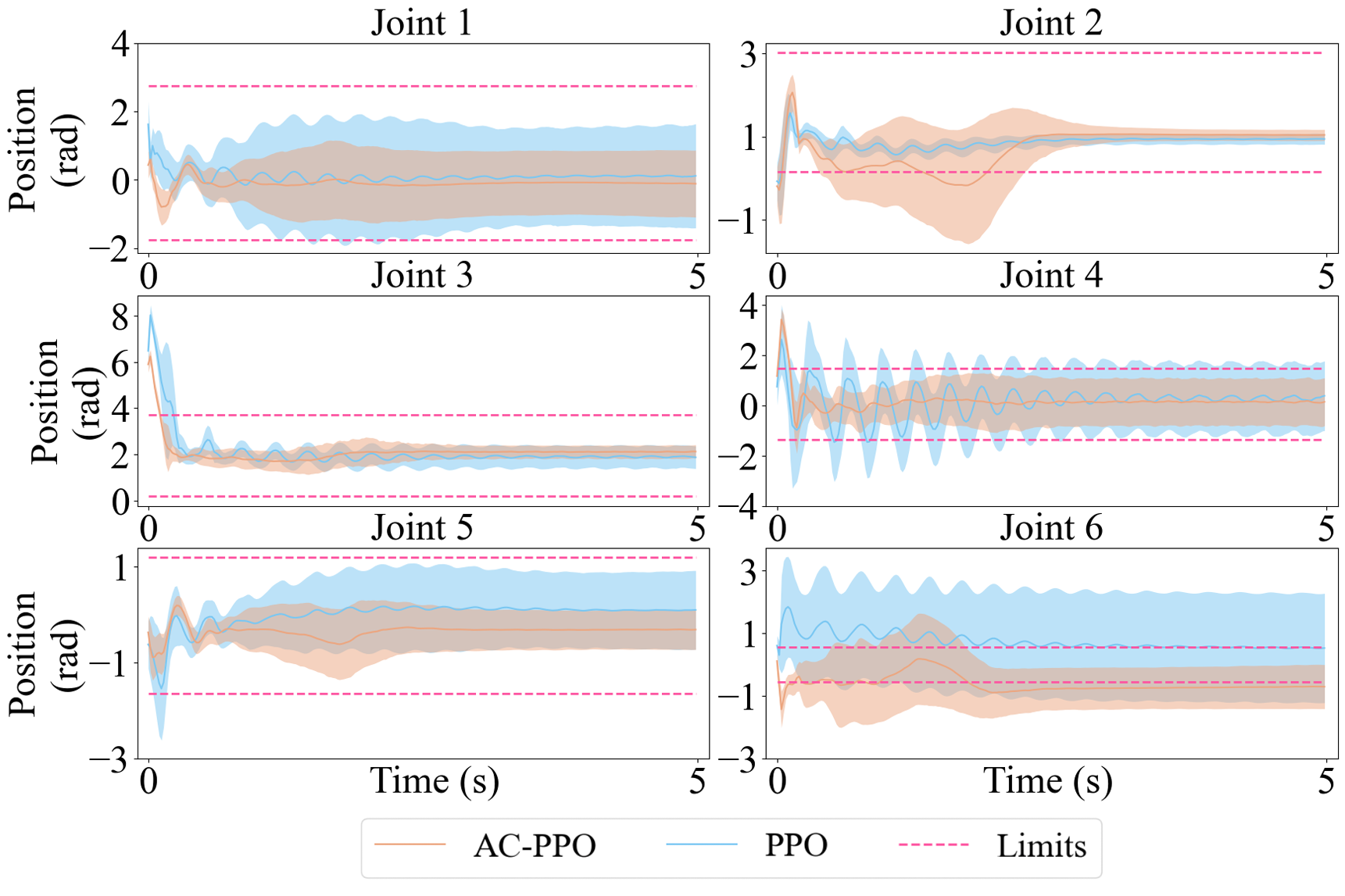}       
    \caption{Effect of the arm-constraints}
    \label{fig:constraints}
\end{figure}

To evaluate the significance of the constraint network, we conducted a comparative analysis with our approach against the reward-only PPO framework, with all training parameters kept identical. As depicted in Fig.~\ref{fig:constraints}, our method notably promotes safer movements and reduces oscillations across all joints of the robotic arm. The actions of the robotic arm are less likely to violate joint limits for most of the joints when compared with the baseline method (only the J6 joint, which connects to the gripper, seems to violate the limitation more than the baseline, but it can be the result of trying to balance the arm centroid and the base centroid). 


\section{Conclusion}
In summary, this paper presents a reinforcement learning framework for the loco-manipulation of wheel-legged robots, enabling them to perform a range of complex manipulation tasks in highly dynamic situations. We emphasize the additional challenges posed by incorporating the arm into the system. Firstly, an AC-PPO is designed to ensure safety and stability in control performance. Secondly, a reward-aware curriculum learning algorithm is proposed to address differences in reward settings between the arm and the base. The structure demonstrates relatively high tracking accuracy in simulation. Finally, we showcase the proficiency of the architecture in the real world by completing basic teleoperation tasks and dynamic manipulation tasks. In the future, we are going to implement and extend the architecture to the multi-agent collaboration tasks.

\balance
\bibliographystyle{IEEEtran}
\bibliography{IEEEexample}

\end{document}